\documentclass[letterpaper, 10 pt, conference]{ieeeconf}  

\IEEEoverridecommandlockouts                              

\usepackage{graphicx}

\usepackage{amsthm}
\usepackage{amssymb}
\usepackage{xspace}
\usepackage{caption}
\usepackage{subcaption}
\usepackage{calrsfs}
\usepackage{bm}
\usepackage[fleqn]{mathtools}
\usepackage{color}
\usepackage{booktabs}
\usepackage{setspace}
\usepackage{float}
\usepackage{siunitx}
\usepackage{multirow}
\usepackage{mdsymbol}
\usepackage{hyperref}
\usepackage{booktabs}%
\usepackage{algorithm}%
\usepackage{algorithmicx}%
\usepackage[noend]{algpseudocode}%
\usepackage{listings}%
\usepackage{hyphenat}
\usepackage{url}
\usepackage{amsmath}
\usepackage[displaymath, mathlines]{lineno}

\newcommand{\BibTeX}{\rm B\kern-.05em{\sc i\kern-.025em b}\kern-.08em\TeX}

\newcommand{\no}[1]{{\; {not} \;}}
\newcommand{\myif}[1]{{\texttt{:-}}}

\DeclareMathAlphabet{\pazocal}{OMS}{zplm}{m}{n}
\theoremstyle{definition}
\newtheorem{assn}{Assumption} 

\newenvironment{frcseries}{\fontfamily{frc}\selectfont}{}

\title{\LARGE \bf
Explainable Online Unsupervised Anomaly Detection for Cyber-Physical Systems via Causal Discovery from Time Series*
}

\author{Daniele Meli$^{1}$
\thanks{*This project has received funding from the Italian Ministry for University and Research, under the PON “Ricerca e Innovazione” 2014-2020" (grant agreement No. 40-G-14702-1).}
\thanks{$^{1}$The author is with Department of Computer Science, University of Verona, Italy
        {\tt\small daniele.meli@univr.it}}%
}

\begin{document}

\maketitle
\thispagestyle{empty}
\pagestyle{empty}

\begin{abstract}

Online unsupervised detection of anomalies is crucial to guarantee the correct operation of cyber-physical systems and the safety of humans interacting with them.
State-of-the-art approaches based on deep learning via neural networks achieve outstanding performance at anomaly recognition, evaluating the discrepancy between a normal model of the system (with no anomalies) and the real-time stream of sensor time series. However, large training data and time are typically required, and explainability is still a challenge to identify the root of the anomaly and implement predictive maintainance. 
In this paper, we use causal discovery to learn a normal causal graph of the system, and we evaluate the persistency of causal links during real-time acquisition of sensor data to promptly detect anomalies. On two benchmark anomaly detection datasets, we show that our method has higher training efficiency, outperforms the accuracy of state-of-the-art neural architectures and correctly identifies the sources of $>10$ different anomalies. The code is at \href{https://github.com/Isla-lab/causal_anomaly_detection}{https://github.com/Isla-lab/causal\_anomaly\_detection}.

\end{abstract}

\section{INTRODUCTION}
Cyber-Physical Systems (CPSs) permeate modern industry and society, from process automation to robotics \cite{chen2017applications}.
It is then fundamental to ensure their continuous service and avoid potential harms to human operators, which may be caused by physical deterioration of mechanical and electronical components, or induced by malware cyber-attacks \cite{alguliyev2018cyber}. 
In this scenario, anomaly detection is crucial for timely identification of deviations from the expected behavior of the CPS \cite{tan2020brief}.
In particular, \emph{Online Unsupervised Anomaly Detection (OUAD)} aims at identifying anomalies during system execution, requiring only a nominal (or normal) model of the system without anomalies, and verifying its adherence to the stream of information coming from the plant \cite{grass2019unsupervised}.
OUAD has been boosted by the recent uptake of deep learning \cite{luo2021deep}, which can infer the nominal model of the system from raw sensor data of recorded executions (typically represented as time series \cite{blazquez2021review}), in most cases outperforming classical machine learning \cite{ruff2021unifying}.
However, neural-based OUAD algorithms typically require significant computational and data resources at training stage, and provide no explanation about the potential sources of the anomaly, hindering, e.g., predictive maintainance \cite{luo2021deep}.
Explainability can be realized by analyzing the learned neural model for individual sensors \cite{zhang2019deep,brigato2021exploiting}, while not reducing the training effort and being subject to the specific choice of neural architecture and  hyperparameters \cite{luo2021deep}. Similarly, post-training explanations can be even more computationally expensive and biased \cite{chernyshov2023explainable}.

In this paper, we propose to use causal discovery \cite{assaad2022survey} to learn a normal temporal causal model of the CPS from time series. Already employed for industrial process prediction in combination with deep learning \cite{menegozzo2021industrial}, causal analysis is more powerful and accurate than mere statistical clustering used for explainable deep learning \cite{wickramasinghe2021explainable}, since it captures the \emph{true causal connections} between variables in the plant. Differently from mixed causal-neural architectures \cite{tang2023causality}, we use state-of-the-art PCMCI algorithm \cite{pcmci} to learn normal causal links, and propose a novel OUAD algorithm to estimate causal correlation during real-time acquisition of sensor data. In this way, we reduce training time and efficiently identify broken causal connections, corresponding to potential sources of anomaly.
We validate our methodology against state-of-the-art neural models, on two benchmark datasets of CPSs: the Secure Water Treatment (SWAT) plant \cite{goh2017dataset} and Pepper social robot by SoftBank Robotics \cite{pandey2018pepper}.

\section{CAUSAL DISCOVERY}\label{sec:background}
We consider a physical process described by a set of $N$ time series $\bm{X} = \{X^j\}_{j=1,\ldots N}$, observed for a finite horizon $T$. We denote as $X^j = [x_1^j, \ldots x_T^j]$ the sequence of observations of variable $X^j$ for $T$ time steps.

The goal of causal discovery is to identify \emph{directed causal links} between variables in $\bm{X}$.
More specifically, causal links are determined according to the measure of \emph{Conditional Mutual Information (CMI)}, which is defined for random variables $X, Y, Z$ as:
\begin{equation*}
    I(X;Y|Z) = \iiint p(x,y,z) \log{\frac{p(x,y|z)}{p(x|z)p(y|z)}} dx dy dz
\end{equation*}
where $p(\cdot|\cdot)$ and $p(\cdot, \cdot)$ denote the conditional and joint probability distributions, respectively.  
From the above definiton, it can be easily shown that variables $X$ and $Y$ are \emph{conditionally independent} under $Z$, denoted as $X \upvDash Y | Z$, iff $I(X;Y|Z) = 0$. In other words, $X$ and $Y$ \emph{have no mutual causal influence}, assuming that $Z$ holds. On the other hand, $X$ and $Y$ \emph{may conditionally depend} on $Z$.
CMI computation depends on the specific analytical assumption for the dependency between time series \cite{runge2018causal}.
With reference to our process $\bm{X}$, we are interested in verifying the following:
\begin{equation}\label{eq:FullCI}
    x^i_{t-\tau} \not\upvDash x^j_t \ | \ \bm{X}_{t-1} \setminus \{x^i_{t-\tau}\}
\end{equation}
i.e., whether the evolution of variable $X^j$ at time $t$ \emph{is caused} by the behavior of $X^i$ at time $t-\tau$ ($\tau > 0$), \emph{conditioning on all other past variables} in $\bm{X}_{t-1}$, representing observations of variables in $\bm{X}$ \emph{up to $t-1$}. In this case, we say that $x^i_{t-\tau}$ is a \emph{causal parent} of $x^j_t$.

Verifying Equation \eqref{eq:FullCI} for all variables and delays of a process is computationally infeasible \cite{runge2018causal}.
More efficient algorithms are available \cite{assaad2022survey}, which are sound and complete under the following assumptions.
\begin{assn}[Causal sufficiency]
Consider $X^i, X^j \in \bm{X}$. Any causal parent of them is in $\bm{X}$.
\end{assn}
In case a variable $Y \not\in \bm{X}$ causes $X^i, X^j$, then the causal analysis of only variables in $\bm{X}$ would highlight a spurious apparent causal correlation between $X^i, X^j$.
\begin{assn}[Causal Markov condition]
If $X^i$ is a causal parent for $X^j$, then the parents of $X^i$ are not parents of $X^j$.
\end{assn}
\begin{assn}[Faithfulness]
The causal model of $\bm{X}$ contains all causal parents following causal Markov condition.
\end{assn}
\begin{assn}[Causal stationarity]
If $x^i_{t-\tau}$ is a causal parent of $x^j_t$, this holds $\forall t \leq T$.
\end{assn}

\noindent
In this paper, we consider the efficient PCMCI algorithm proposed by \cite{pcmci}, which has been successfully applied to complex real-world datasets \cite{runge2019inferring}.
PCMCI has two steps:
\begin{itemize}
    \item the PC algorithm \cite{schmidt2018order}, which iteratively builds the set of causal parents of $x^j_t$, $\pazocal{P}(x^j_t)$, initialized to $\left\{\bm{X}_{t-1}, \ldots, \bm{X}_{t-\tau_{max}}\right\}$, $\tau_{max}$ being the user-defined maximum delay. Specifically, at each step $p$, $\forall x^i_{t-\tau} \in \pazocal{P}(x^j_t)$ we check $x^i_{t-\tau} \not\upvDash x^j_t \ | \ \pazocal{S}$,
    where $\pazocal{S}$ is the subset of first $p$ parents in $\pazocal{P}(x^j_t) \setminus \left\{x^i_{t-\tau}\right\}$, sorted by the highest CMI value. 
    If the above holds, $x^i_{t-\tau}$ is removed from the set of causal parents;
    \item the Momentary Conditional Independence (MCI) test, to prune a spurious parent $x^i_{t-\tau}$ if the following holds:
    \begin{equation*}
        x^i_{t-\tau} \not\upvDash x^j_t \ | \ \pazocal{P}(x^j_t) \setminus \left\{x^i_{t-\tau}\right\}, \pazocal{P}(x^i_{t-\tau})
    \end{equation*}
    In this way, the dependency of parent $x^i_{t-\tau}$ from its causal parents $\left(\pazocal{P}(x^i_{t-\tau})\right)$ is excluded when evaluating its causal effect on $x^j_t$.
\end{itemize}

\section{CAUSAL DISCOVERY FOR OUAD}
\begin{algorithm}[t]
    \caption{Causal discovery for OUAD}\label{alg:ouad}
    \begin{algorithmic}[1]
        \State \textbf{Input}: $\bm{X}^n, \Bar{e}$
        \State \textbf{Output}: Broken links $\pazocal{L}$
        \State $t_s = \frac{1}{10\max{\Gamma}}$ \textit{\% Sampling at 5$\times$ Nyquist frequency}
        \State $\bm{X}^n = \texttt{SUBSAMPLE}(\bm{X}^n, t_s)$
        \State \textit{\% Remove nearly constant time series from the dataset}
        \For{$X^j \in \bm{X^n}$}
        \If{$\texttt{MEAN}(X^j) < 0.01\cdot \texttt{STD}(X^j)$}
        \State $\bm{X}^n.\texttt{REMOVE}(X^j)$
        \EndIf
        \EndFor
        \State $\tau_{max} = \frac{1}{t_s\cdot \texttt{MEAN}(\Gamma)}$ \textit{\% Empirically set the maximum delay to the ratio between the average period of frequency components in the dataset and the sampling rate}
        \State $\pazocal{C}^n \gets \texttt{PCMCI}(\bm{X}^n, \tau_{max})$
        \ForAll{$c_{ij\tau} \in \pazocal{C}^n$}
        \If{$c_{ij\tau} < \texttt{MEAN}(\pazocal{C}^n)$}
        \State $\pazocal{C}^n.\texttt{REMOVE}(c_{ij\tau})$
        \EndIf
        \EndFor
        \newline
        \State \textit{\% Online anomaly detection}
        \State $t=1, \pazocal{L} = \emptyset$
        \For{$t \leq T^a$} \textit{\% Time horizon of $\bm{X}^a$}
        \State Collect $\bm{X}^a_t$
        \State \textit{\% Analyze only at multiples of the subsampling rate}
        \If{$t \bmod t_s = 0$}
        \ForAll{$X^j \in \bm{X}^a_t$}
        \State $\bm{X} = \{X^j\}$
        \ForAll{$X^i \in \bm{X}^a_t$}
        \ForAll{$\tau < \tau_{max}$}
        \If{$c^n_{ij\tau} \neq 0$}
        \State $\bm{X}.\texttt{APPEND}(X^i)$
        \State \textbf{break}
        \EndIf
        \EndFor
        \EndFor
        \State $\pazocal{C}^a_t = \texttt{LEAST\_SQUARES}(\bm{X})$
        \ForAll{$c^{at}_{ij\tau} \in \pazocal{C}^a_t$}
        \If{$\left| c^{at}_{ij\tau} - c^n_{ij\tau} \right| > \Bar{e}$}
        \State $\pazocal{L}.\texttt{APPEND}(\langle i,j \rangle)$
        \EndIf
        \EndFor
        \If{$\pazocal{L} \neq \emptyset$}
        \Return $\pazocal{L}$
        \EndIf
        \EndFor
        \EndIf
        \EndFor
        \Return $\pazocal{L}$
    \end{algorithmic}
\end{algorithm}

We assume to have two datasets of $N$ time series observed over a horizon $T$: a dataset $\bm{X}^n$ representing the evolution of the system without anomalies (\emph{normal}), and a dataset $\bm{X}^a$ of anomalous evolution.
The goal of OUAD is to identify a mathematical model $\pazocal{M}$ approximating the evolution of $\bm{X}^n$, and then to detect anomalies \emph{online} as $\bm{X}^a$ is being collected, by only analyzing its discrepancy with respect to $\pazocal{M}$.

In this paper, $\pazocal{M}$ is a \emph{causal model} detected with any causal discovery algorithm.
As outlined in Algorithm \ref{alg:ouad}, we preliminarily filter time series in $\bm{X}^n$ via subsampling according to their main frequency components (Lines 3-4), and remove nearly constant signals to focus only on relevant temporal variations for causal discovery (Lines 6-8). We then apply PCMCI\footnote{We use the implementation of PCMCI available at \href{https://github.com/jakobrunge/tigramite.git}{https://github.com/jakobrunge/tigramite.git}.} to $\bm{X}^n$ (Line 10).
More specifically, to measure CMI we assume linear dependency between variables and their parents. Hence, computing Equation \eqref{eq:FullCI} resorts to finding Pearson correlation coefficient $c^n_{i,j,\tau}$ between $X^j$ and its parent $X^i$ with delay $\tau$. We denote the matrix of coefficients as $\pazocal{C}^n \in \mathbb{R}^{N\times N\times \tau_{max}}$. Under the common assumption of Gaussian error distribution \cite{runge2018causal}, the CMI test is considered valid only when the p-value is below $0.05$. The linear dependency assumption can be relaxed, and more complex CMI tests can be integrated in PCMCI, e.g., based on parametric regression or k-nearest neighbors \cite{runge2018conditional}. From our experiments, linearity is a sufficient assumption to capture main aspects of the causal model underlying data. Furthermore, it is more computationally convenient for the online phase of OUAD, as explained below.

Once the causal model $\pazocal{M}$ is acquired, we empirically filter relevant causal links by selecting only the ones with coefficient $c^n_{ij\tau}$ higher than the average over the whole $\pazocal{C}^n$ (Lines 11-13).
Then, in the \emph{online phase}, we progressively collect $\bm{X}^a_t$ over time (Line 17). Analyzing each time series $X^j \in \bm{X}^a_t$, from $\pazocal{M}$ its causal connections to other variables should be represented as:
\begin{equation*}
    x^j_t = \sum_{i=1}^N \sum_{\tau=1}^{\tau_{max}} c^n_{ij\tau} x^i_{t-\tau}
\end{equation*}
We then compute Pearson coefficients between $x^j_t$ and all lagged variables such that $c^n_{ij\tau} \neq 0$ (Lines 20-27).

This results in a new matrix of coefficients $\pazocal{C}^a_t = \{c^{at}_{ij\tau}\}$, acquired from the stream of time series in $\bm{X}^a_t$.
We then compare $\pazocal{C}^a_t$ vs. $\pazocal{C}^n$, computing the error $\left| c^{at}_{ij\tau} - c^n_{ij\tau} \right|$. Whenever this error is greater than a predefined threshold $\Bar{e}$, it means that the causal link between $X^j$ and $X^i$ with delay $\tau$ is broken, and an anomaly alarm is raised, returning the set of broken causal links (Lines 28-31).
In this paper, we empirically choose a different threshold\footnote{The threshold value was tuned to maximize OUAD accuracy.} $\Bar{e} = \Bar{e}_{ij\tau} \ \forall j,i=1,\ldots,N, \tau=1,\ldots,\tau_{max}$:
\begin{equation}\label{eq:threshold}
    \Bar{e}_{ij\tau} = \sqrt{\sum_{t=1}^{T^n} \left(c^n_{ij\tau} - \hat{c}^{nt}_{ij\tau}\right)^2}
\end{equation}
where $T^n$ is the time horizon of the normal dataset, and $\hat{c}^{nt}_{ij\tau}$ is the coefficient representing the causal relationship between $x^j_{t}, x^i_{t-\tau} \in \bm{X}^n_t$, estimated via least squares approximation on $\bm{X}^n_t$.
In other words, we preliminarily compute coefficients representing causal relations observed in the normal dataset at progressive time horizon, similarly to the online procedure adopted for computing $\pazocal{C}^a_t$. We then estimate the alarm threshold for each coefficient as the 2-norm of the errors observed in the normal dataset (where we expect errors $\approx 0$).

\noindent
With our methodology, we are able not only to detect anomalies in real time, but we can also gather information about \emph{the causal root of the anomaly}, by analyzing the error on single Pearson coefficients (corresponding to a specific causal link between two signals). Moreover, the linear assumption on the causal model significantly simplifies online coefficient computation.

\section{DATASETS}
\begin{figure}
    \centering
    \begin{subfigure}{0.3\textwidth}
    \includegraphics[scale=0.6]{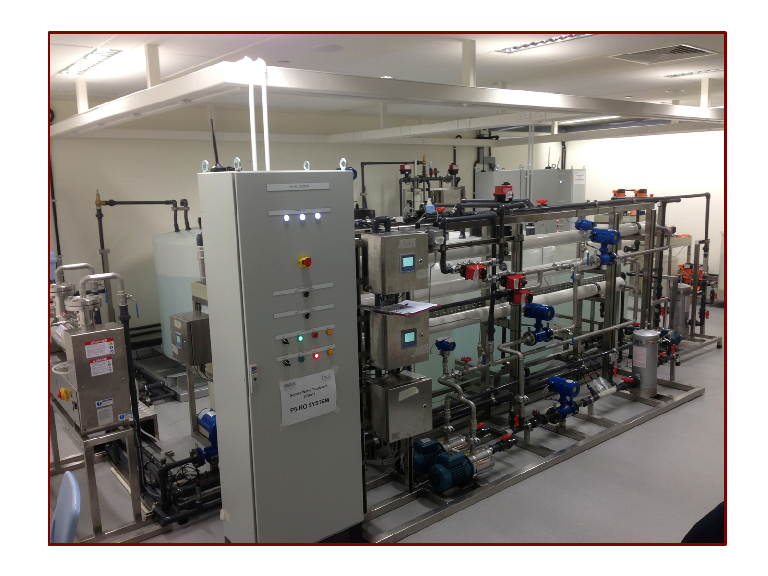}
    \caption{\label{fig:swat}}
    \end{subfigure}
    \begin{subfigure}{0.1\textwidth}
    \includegraphics[scale=0.05]{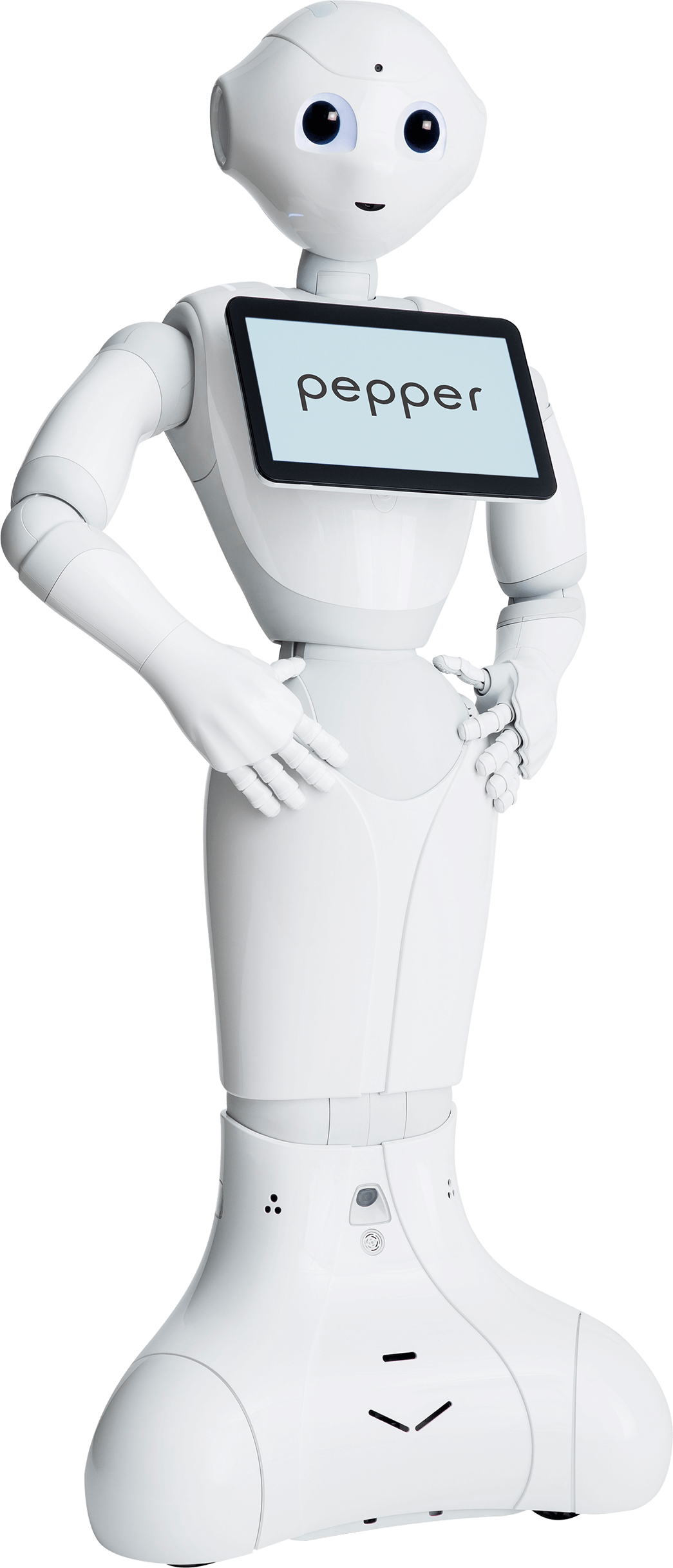}
    \caption{\label{fig:pepper}}
    \end{subfigure}\\
    \begin{subfigure}{0.5\textwidth}
    \centering
    \includegraphics[scale=0.27]{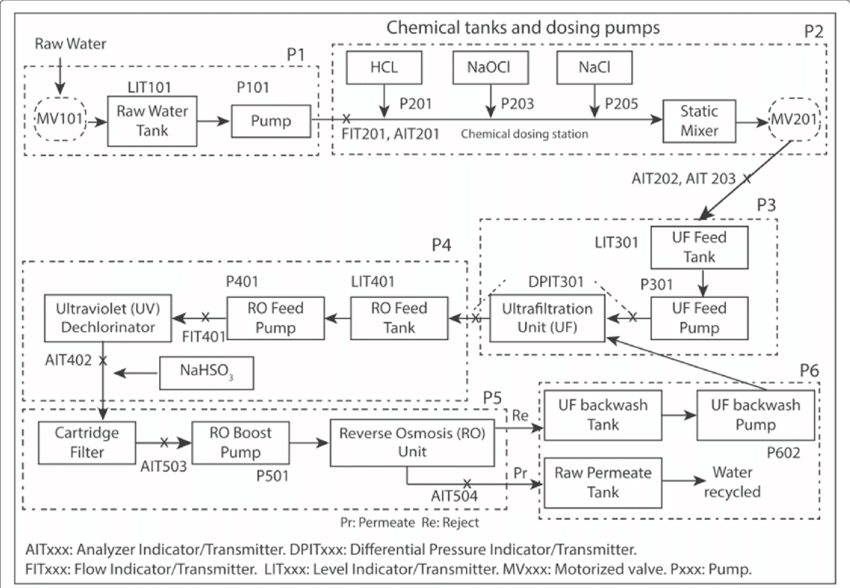}
    \caption{\label{fig:swat_scheme}}
    \end{subfigure}
    \caption{The two testing scenarios: a) SWAT; b) Pepper social robot; c) a functional scheme of SWAT stages P1-P6.}
    \label{fig:setups}
\end{figure}
We now briefly describe the datasets used for validating our methodology, namely, the SWAT industrial system and the social robot Pepper.
For full details about them, we refer the reader respectively to \cite{goh2017dataset} and \href{https://sites.google.com/diag.uniroma1.it/robsec-data}{https://sites.google.com/diag.uniroma1.it/robsec-data}.

\subsection{SWAT}
The SWAT dataset contains signal recordings from a realistic small-scale replication of an industrial plant for water filtration, developed at the University of Singapore (Figure \ref{fig:swat}). 
It is a 6-stage process, characterized by 51 physical variables representing actuators (e.g., pumps and valves) and sensors (e.g., pressure and flow meters).

The dataset\footnote{\href{https://itrust.sutd.edu.sg/itrust-labs\_datasets/dataset\_info/}{https://itrust.sutd.edu.sg/itrust-labs\_datasets/dataset\_info/}, 2015 version.} considered in this paper consists of both normal (i.e., anomaly-free) and anomalous acquisitions.
The normal dataset records 7 days of operation.
The anomalous dataset records 4 days of operation presenting sequences of different anomalies, generated from cyberphysical attacks at different stages of the process.
Due to the many different anomaly classes, the SWAT dataset is a popular benchmark for the analysis of cyberphysical systems, as well as cybersecurity \cite{conti2021survey} and anomaly detection in general \cite{luo2021deep, castellini2023adversarial}.

From the preliminary filtering procedure described at Lines 3-4 of Algorithm \ref{alg:ouad}, it turns out that an adequate sampling time for SWAT dataset is $\approx\SI{3}{min}$; therefore, in this paper we only evaluate attacks lasting longer than this time lapse, as reported in Table \ref{tab:swat} (variable names are in the form \emph{N-XYZ}, where \emph{X} indicates a stage of the process).
\begin{table*}
    \centering
    \caption{Attack scenarios in SWAT considered in this paper, with IDs from the original dataset.}    
    \resizebox{0.7\textwidth}{!}{%
    \begin{tabular}{c|c}
        \textbf{ID} & \textbf{Description}\\
        \toprule
        8 & Value of DPIT-301 (pressure sensor) increased, tank 401 level increases, tank 301 decreases\\
        17 & MV-303 (valve actuator) closed, stage 3 stops\\
        21 & MV-101 open, modified LIT-101 (level actuator)\\
        23 & Value of DPIT-301 increased, MV-302 open, P-602 (pump actuator) open\\
        25 & LIT-401 modified, P-402 open\\
        26 & P-101 activated, LIT-301 is modified\\
        27 & LIT-401 is modified\\
        28 & P-302 is closed\\
        30 & P-101 and MV-101 are activated, LIT-301 modified, P-102 activates\\
    \end{tabular}}
    \label{tab:swat}
\end{table*}

\subsection{Pepper}
\label{sec:pepper}
Pepper (Figure \ref{fig:pepper}) is a humanoid robot designed by SoftBank Robotics for human-robot interaction and social assistance \cite{pandey2018pepper}.
The dataset consists of 256 sensor readings, including the kinematic state of joints and wheels, accelerometers, LED and laser configurations for perception and communication.
Specifically, 14000 time points are collected under nominal behavior, while 4244 time points are recorded under the effect of 3 different cyberphysical attacks:
\begin{itemize}
    \item \emph{LED}, i.e.,  pseudo-random remote activation of LEDs;
    \item \emph{Joints}, i.e., remote control of joints;
    \item \emph{Weights}, i.e., remote control of wheels.
\end{itemize}
Pepper dataset is an established benchmark for robotic anomaly detection \cite{olivato2019comparative,brigato2021exploiting}, particularly challenging for the high number of variables.

\section{RESULTS}
We validate Algorithm \ref{alg:ouad} on SWAT and Pepper datasets. 
We first analyze the output and performance of PCMCI on the normal dataset (resulting in the normal causal description of the process).
We then make a \emph{quantitative assessment} of the precision, recall and F1-score achieved by our methodology in detecting different anomalies for SWAT (Table \ref{tab:swat}) and Pepper (Section \ref{sec:pepper}). We compare against 3 state-of-the-art algorithms for OUAD based on neural networks, following the implementation in \cite{brigato2021exploiting}: 
\begin{itemize}
    \item \emph{RNN}: a recurrent neural network composed of a gated recurrent unit and a 64-node hidden layer, plus a dropout layer to the output;
    \item \emph{TCN}: a temporal convolutional network with 3 30-node hidden layers and 3-size kernel;
    \item \emph{TCE}: a temporal convolutional autoencoder, which analyzes the time series of each variable $X^j$ converted to a 2D image, where pixel values depend on $\Dot{X}^j, \Ddot{X}^j$. For SWAT dataset, TCE consists of 6 convolutional layers (respectively, 32, 16, 8, 16, 32, 3 nodes with 6-size kernels), while for Pepper it consists of 8 convolutional layers (respectively, 100, 50, 30, 10, 30, 50, 100, 3 nodes with 6-size kernels). Temporal autoencoders are a state of the art for time series anomaly detection \cite{li2023deep}.
\end{itemize} 
We finally perform a \emph{qualitative analysis} of the detected broken causal links for different anomalies, to show the explainability of our methodology.

All experiments are run on a PC with Intel i9-13900HX 24-core CPU (\SI{2.2}{MHz}).

\subsection{Normal causal models}\label{sec:res_normal_model}
\begin{figure*}
    \centering
    \includegraphics[scale=0.4]{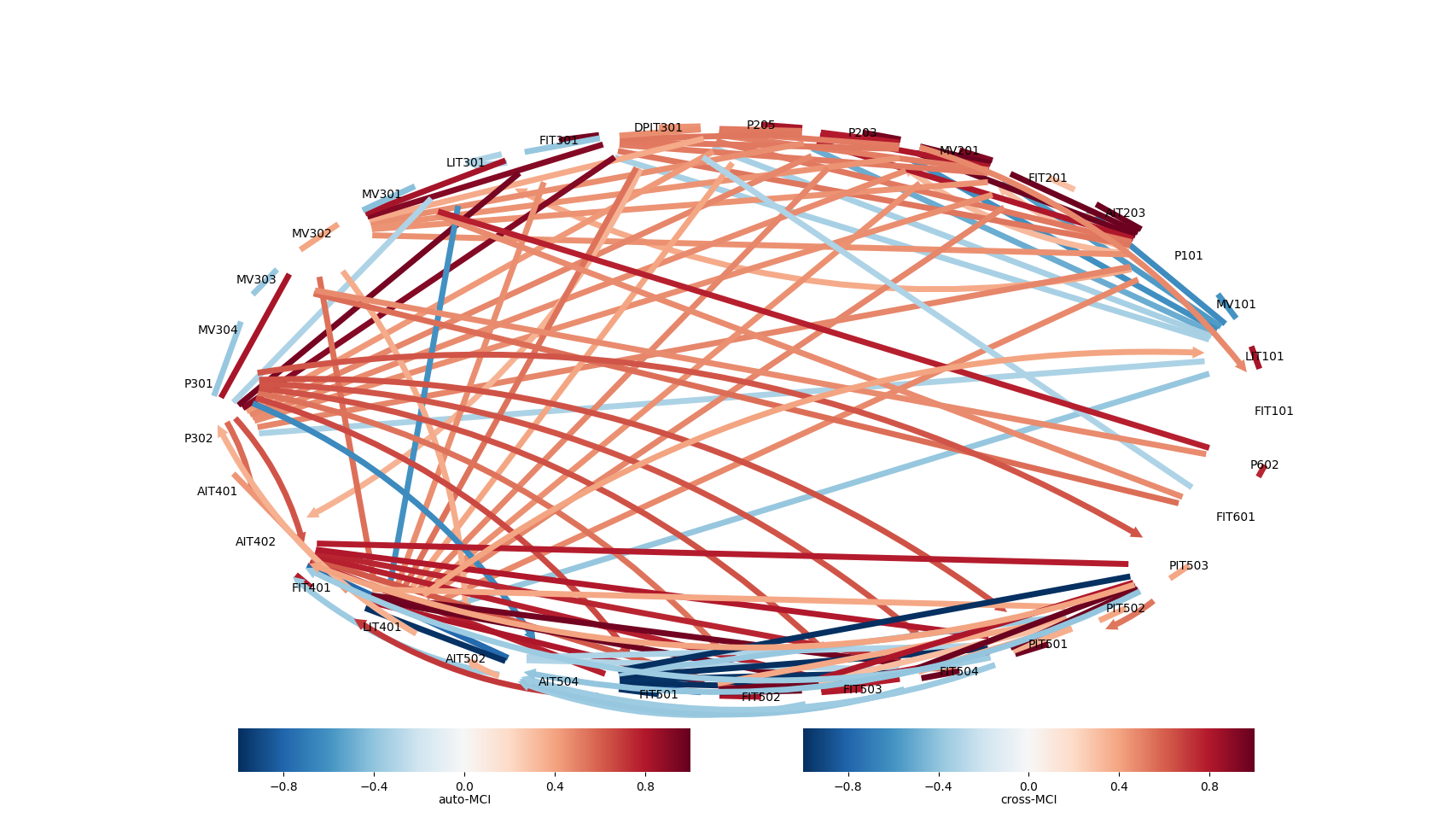}
    \caption{Normal causal graph for SWAT. \emph{auto-MCI} is for autoregressive dependency, \emph{cross-MCI} is for cross-variable dependency. Only links with $|\text{MCI}|>0.3$ are shown for ease of visualization.}
    \label{fig:causal_model}
\end{figure*}
Following the methodology described in Lines 3-13 of Algorithm \ref{alg:ouad}, we use PCMCI to identify a normal causal model for Pepper within $\approx\SI{320}{s}$, and for SWAT within $\approx\SI{25}{s}$. As a comparison, training the neural models takes, on average, $\approx\SI{500}{s}$ for Pepper and $\approx\SI{220}{s}$ for SWAT, proving the reduced computational time required by causal analysis.

In SWAT dataset, 33 out of 51 variables are included in the causal model.
For a better understanding of the causal model, in Figure \ref{fig:causal_model} we represent it as a causal graphs, where nodes represent process variables and arrow links denote directed connections from causal parents to effects (temporal delays $\tau$'s are omitted for the ease of visualization).
We notice that variables of a generic stage $X$ are connected and caused by variables at stage $X-1$, which represents the sequence between stages depicted in Figure \ref{fig:swat_scheme}. It is also interesting that variables of stage 6 are identified as causal parents of several variables at stage 3. This correctly reflects the connection shown in Figure \ref{fig:swat_scheme} between stages 6 and 3, and particularly the dependency of the ultrafiltration (UF) unit (connecting stages 3 and 4) from the backwash pump P-602.
Finally, P-301 UF pump actuator at stage 3 affects variables at stage 5, which involves reverse osmosis started at stage 4.

In Pepper dataset, 243 out of 255 variables are included in the causal model. Excluded variables are related to head touch sensors and bumpers, which are indeed rarely activated during the normal operation of Pepper robot. On the other hand, variables describing the kinematic state of the robot (joint and Cartesian positions, velocities, accelerations and jerks), laser and LEDs are correctly included in the normal causal model.
Due to the large number of variables in Pepper causal model, we cannot visualize its graph.

\subsection{Quantitative analysis}
\begin{table}
    \centering
    \caption{Quantitative results of OUAD in percentages (\underline{\textbf{best}}). Following \cite{brigato2021exploiting}, RNN and TCN for SWAT are preceded by principal component filtration for better results.}    
    {%
    \begin{tabular}{c|c|c|c|c|c|c}
    \multirow{2}{*}{} & \multicolumn{3}{c|}{SWAT} & \multicolumn{3}{c}{Pepper} \\
    \cline{2-7}
    & \emph{Pr} & \emph{Rec} & \emph{F1}& \emph{Pr} & \emph{Rec} & \emph{F1} \\
    \toprule
    \textbf{Ours} & \underline{\textbf{100}} & \underline{\textbf{100}} & \underline{\textbf{100}} & 66 & 92 & 77 \\
    \textbf{RNN} & 96 & 63 & 76 & 84 & 79 & 82 \\
    \textbf{TCN} & \underline{\textbf{100}} & 59 & 74 & \underline{\textbf{89}} & 73 & 80 \\
    \textbf{TCE} & 99 & 58 & 73 & 86 & \underline{\textbf{95}} & 90 \\
    \end{tabular}}
    \label{tab:results}
\end{table}
Table \ref{tab:results} reports Precision ($Pr = \frac{TP}{TP+FP}$), Recall ($Rec = \frac{TP}{TP+FN}$) and F1-score ($F1 = \frac{2PrRec}{Pr+Rec}$) of OUAD with different algorithms and datasets, cumulating over all anomaly classes. \emph{TP} are true positives, i.e., correctly identified anomalies; \emph{FP} are false positives, i.e., anomalies mistakenly identified on normal data points; \emph{FN} are false negatives, i.e., missed anomalies.
For better statistical relevance, we run PCMCI on $70\%$ of the normal dataset $\bm{X}^n$, thus including the left $30\%$ in $\bm{X}^a$ when executing Lines 16-31 of Algorithm \ref{alg:ouad}. In this way, we learn coefficients $\{c^n_{ij\tau}\}$ (Line 10) \emph{only from a part} of the normal dataset, which are then used at Line 29 to raise an anomalous alarm. Otherwise, replacing $c^{at}_{ij\tau}$ with $\hat{c}^{nt}_{ij\tau}$ (i.e., online computed causal coefficients $\hat{\pazocal{C}}^n$ from $\bm{X}^n$) at Line 29 $\forall i,j \leq N, \tau \leq \tau_{max}, t\leq T$ it would clearly hold:
\begin{equation*}
    \left| \hat{c}^{nt}_{ij\tau} - c^n_{ij\tau} \right| \leq\ \Bar{e} = \sqrt{\sum_{t=1}^{T^n} \left(c^n_{ij\tau} - \hat{c}^{nt}_{ij\tau}\right)^2}
\end{equation*}
thus, \emph{FP}=0 and $Pr = 100\%$ trivially.
Our methodology achieves 100\% scores on SWAT. On Pepper, it does not have the best performance; however, the recall (92\%), measuring the rate of correct anomalies vs missed ones, is comparable to the best one by TCE (95\%). The average and standard deviation of the computational time required by Algorithm \ref{alg:ouad} at each online iteration (Lines 17-30) is $1 \pm 4$ ms.

\subsection{Anomaly explanations}
We now evaluate the explainability of our methodology for OUAD.
For each dataset and anomaly $\bm{X}^a$, we sort variables $X^a_j$'s according to the highest value for
\begin{equation}\label{eq:explain_criterion}
    \sqrt{\sum_{i=1}^N\sum_{t=1}^{T^a} \left(c^n_{ij\tau} - c^{at}_{ij\tau}\right)^2}
\end{equation}
which represents the 2-norm of the coefficient errors for $X^a_j$'s causal children over the horizon $T^a$ of $\bm{X}^a$.
In this way, we identify variables of the process whose causal links are more severely broken during the anomaly, which could provide a better insight on its source.

For SWAT dataset, variables corresponding to stages affected by anomalies in Table \ref{tab:swat} are always correctly identified by our algorithm in the top-$10\%$ list (3/33 variables) according to Equation \eqref{eq:explain_criterion}.
For instance\footnote{For compactness, we only describe some representative anomalies.}, in case of anomaly 8, we identify an anomaly on FIT-401 (flow sensor for tank 4). Indeed, as described in Table \ref{tab:swat}, the anomaly on DPIT-301 sensor alters the behavior of tank 4. In particular, the causal connection between FIT-401 and the pressure sensors at stage 5 is broken. In fact, stages 4 and 5 are tightly connected, both concerning the reverse osmosis phase to remove impurities from water.
In case of anomaly 26, affecting both stages 1 and 3, our algorithm captures an alteration in the dependency between stage 4 (flow sensor FIT-401 for tank 4) and actuator P-301 at stage 3, which pumps water to the reverse osmosis unit and is affected by the tank level sensor LIT-301 (directly manipulated in this anomaly, from Table \ref{tab:swat}). Moreover, the contemporaneous attack on pump actuator P-101 and LIT-301 alters the causal relation between P-101 and the reverse osmosis stages 4-5 (particularly pressure sensor PIT-503 and flow sensor FIT-401), which depend on stage 3. The same behavior is observed in anomaly 30, which is similar to anomaly 26.
Finally, in case of anomaly 27 (at stage 4), broken links involve variables of interconnected stages 4 and 5. In particular, the variable with the highest value for Equation \eqref{eq:explain_criterion} is flow sensor FIT-401, directly connected to the attacked level sensor LIT-401.

For Pepper dataset, the \emph{LED} anomaly is due to random activation of LEDs. These are used by the robot to communicate to humans, hence it is hard to detect the correct source of the anomaly by inspecting the causal model of the system.
However, it is interesting to note that the variables with the highest value for Equation \eqref{eq:explain_criterion} are related to the touchpad pressure sensors. As stated in SoftBank's guide\footnote{\href{https://qisdk.softbankrobotics.com/sdk/doc/pepper-sdk/ch6\_ux/chap4.html}{https://qisdk.softbankrobotics.com/sdk/doc/pepper-sdk/ch6\_ux/chap4.html}}, a standard good practice in programming Pepper is to switch off LEDs while the user is interacting with the tablet, in order not to mislead him with other signals. Hence, an anomaly detected on the touchpad sensors may indicate that the LEDs wrongly activated during interaction with the human.
For \emph{Joints} anomaly, the variables with the highest value for Equation \eqref{eq:explain_criterion} are related to the temperature of the knee's pitch angle, due to the pseudo-random joint manipulation. Moreover, considering the top-$10\%$ variables (24/243) ranked by Equation \eqref{eq:explain_criterion}, nearly half of them (i.e., 11) refer to robot joints.
On the other hand, for \emph{Wheels} anomaly, only 2/24 variables are related to joints, and first 4 variables refer to stiffness and temperature of attacked wheels. 

\section{CONCLUSION}
In this paper, we proposed PCMCI for causal discovery of the normal model of execution of a CPS, and a novel OUAD algorithm based on the real-time evaluation of the CMI coefficients associated with causal links. On two benchmark CPS datasets, industrial SWAT and Pepper robot, our algorithm achieves higher or comparable recall than state-of-the-art TCE (100\% vs. 76\% on SWAT, 92\% vs. 95\% on Pepper), thus missing the least of anomalies. Moreover, causal discovery requires less training time and allows to explain the sources of multiple anomalies.

In future works, we will validate on other relevant OUAD datasets, e.g., from the robotics domain \cite{brockmann2023voraus}. Moreover, we will investigate more complex (e.g., non-linear or non-analytical) causal structures, and perform a  more extended hyperparameter analysis (e.g., on alarm threshold $\Bar{e}$).

\bibliographystyle{ieeetr}
\bibliography{biblio.bib}

\end{document}